\documentclass{turing2012}
\usepackage{times}
\usepackage{graphicx, color, import}
\usepackage{latexsym}
\usepackage{amsmath, amsfonts, mathrsfs}

\newcommand{\e}{\varepsilon}

\begin{document}
\newtheorem{dfn}{Definition}
\newtheorem{lem}[dfn]{Lemma}
\newtheorem{prop}[dfn]{Proposition}
\newtheorem{thm}[dfn]{Theorem}
\newtheorem{cor}[dfn]{Corollary}
\newtheorem{fact}[dfn]{Fact}
\newtheorem {exa}[dfn]{Example}
\newcommand{\vsp}{\vspace{6mm}}
\newcommand{\baralp}[2]{\mbox{$\bar{\alpha}_{#1}^{( #2 )}$}}
\newcommand{\alp}[2]{\mbox{$\alpha_{#1}^{( #2 )}$}}
\newcommand{\vex}[1]{\vec{ #1}}
\newcommand{\V}[1]{\mbox{$V^{(#1)}$}}
\newcommand{\VK}[2]{\mbox{$V^{(#1)}(#2)$}}
\newcommand{\SL}[1]{\mbox{$SL^{(#1)}$}}
\newcommand{\LA}[1]{\mbox{$L^{(#1)}$}}
\newcommand{\fn}{\mbox{$f^{(n)}$}}
\newcommand{\pr}[1]{\mbox{$Prob^{(#1)}$}}

\title{The Utility of Hedged Assertions in the Emergence of Shared Categorical Labels}

\author{Martha Lewis \and Jonathan Lawry \institute{University of Bristol, England, email: martha.lewis@bristol.ac.uk, j.lawry@bristol.ac.uk} }

\maketitle
\bibliographystyle{AISB}

\begin{abstract}
We investigate the emergence of shared concepts in a community of language users using a multi-agent simulation. We extend results showing that negated assertions are of use in developing shared categories, to include assertions modified by linguistic hedges. Results show that using hedged assertions positively affects the emergence of shared categories in two distinct ways. Firstly, using contraction hedges like `very' gives better convergence over time. Secondly, using expansion hedges such as `quite' reduces concept overlap. However, both these improvements come at a cost of slower speed of development.
\end{abstract}

\section{INTRODUCTION}
An evolutionary approach to semantics enables the development in robots and autonomous agents of flexible, mutable concepts that could be learnt through interaction and can change over time \cite{steels}. This approach is investigated by Eyre and Lawry in \cite{ettie}, in which they develop a model of language evolution based in the label semantics framework. They show that using a mixture of positive and negated assertions enables the development of languages that are both shared, and discriminate effectively between elements within the environment. We extend this work to include assertions modified by the words `very' and `quite', and show that doing so improves performance in two ways. Use of the hedge `very' improves levels of convergence attained. Using the hedge `quite' reduces the amount of overlap within an agent's label set. We describe in detail the theoretical approach to concepts taken and linguistic hedges in the remainder of this section. Section \ref{sec:method} gives details of the mathematical and computational model used in the simulations. Section \ref{sec:results} gives results of the simulations which are discussed in section \ref{sec:discussion}. Lastly, section \ref{sec:conc} gives conclusions and further avenues of research.

\subsection{A representation model for concepts}
We model concepts within the label semantics framework \cite{lawry2004, lawry2009}, combined with prototype theory \cite{rosch} and the conceptual spaces model of concepts \cite{gard2004}. Prototype theory offers an alternative to the classical theory of concepts, basing categorization on proximity to a prototype. This approach is based on experimental results where human subjects were found to view membership in a concept as a matter of degree, with some objects having higher membership than others \cite{rosch}. Fuzzy set theory \cite{zadeh1965}, in which an object $x$ has a graded membership $\mu_L(x)$ in a concept $L$, was proposed as a formalism for prototype theory. However, numerous objections to its suitability have been made \cite{osh1981, smith, kp, hamp, hamp1987}. 

\begin{figure}[h]
\centering
\includegraphics[width = 0.3\textwidth]{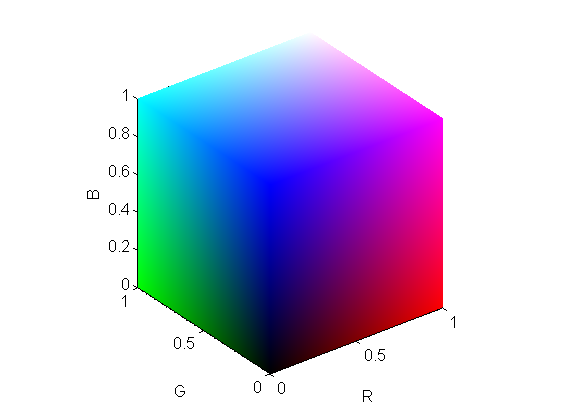}
\caption{The RGB cube represents colours in three dimensions of Red, Green and Blue. A colour concept such as `purple' can be represented as a region of this conceptual space.}
\label{fig:rgbcube}
\end{figure}

Conceptual spaces theory renders concepts as convex regions of a \emph{conceptual space} - a geometrical structure with quality dimensions and a distance metric. Examples are: the RGB colour cube, pictured in figure \ref{fig:rgbcube}; physical dimensions of height, breadth and depth; or the taste tetrahedron. Since concepts are convex regions of such spaces, the centroid of such a region can naturally be viewed as the prototype of the concept.

Label semantics \cite{lawry2004} is a random set approach to concepts which quantifies an agent's uncertainty about the extent of application of a concept. We refer to this as \emph{subjective uncertainty} \cite{lawry2009} to emphasise that it concerns the definition of concepts and categories, in contrast to stochastic uncertainty which concerns the state of the world. Lawry and Tang \cite{lawry2009} combine the label semantics approach with conceptual spaces and prototype theory, to give a formalisation of concepts as based on a prototype and a threshold, located in a conceptual space. 

\begin{figure}[h]
\centering
\includegraphics[width = 0.3\textwidth]{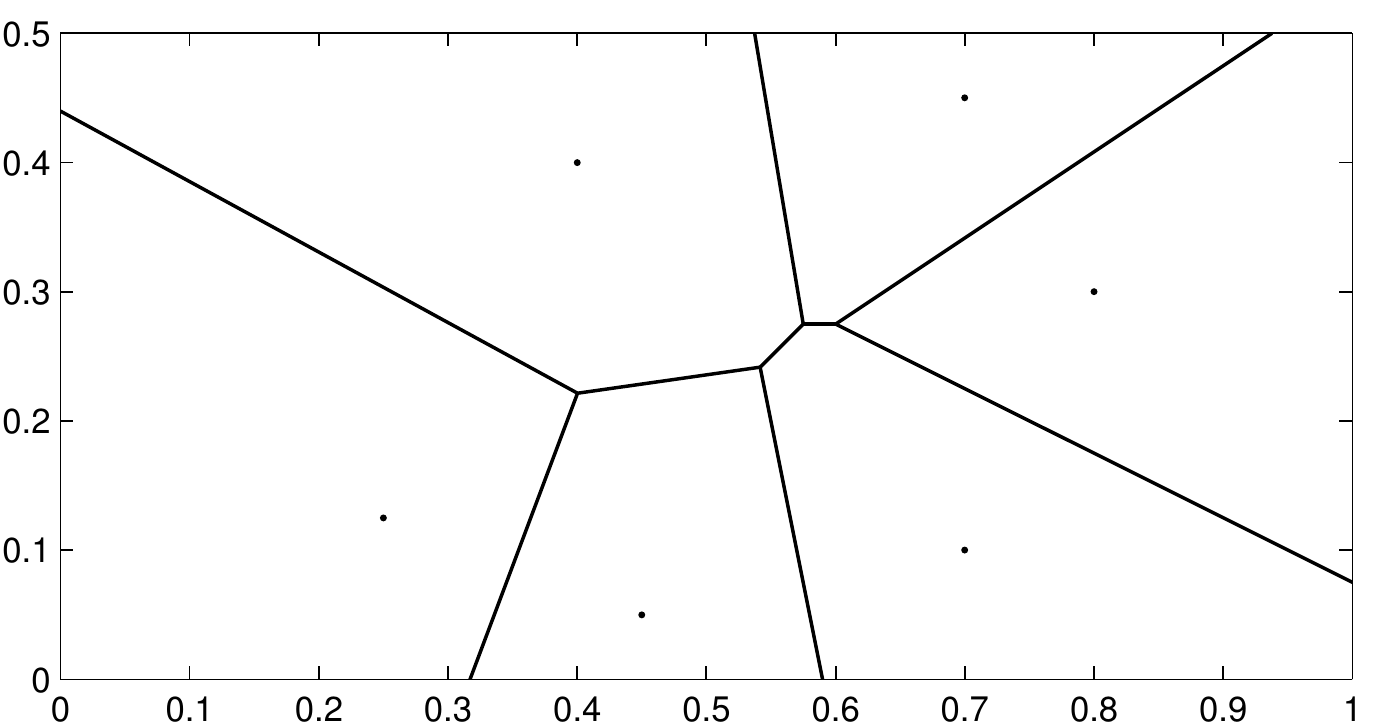}
\caption{Conceptual space divided into concepts according to a Voronoi tessellation around prototypes. Each part of the space corresponds to exactly one concept.}
\label{fig:vplot}
\end{figure}

\begin{figure}[h]
\centering
\includegraphics[width = 0.3\textwidth]{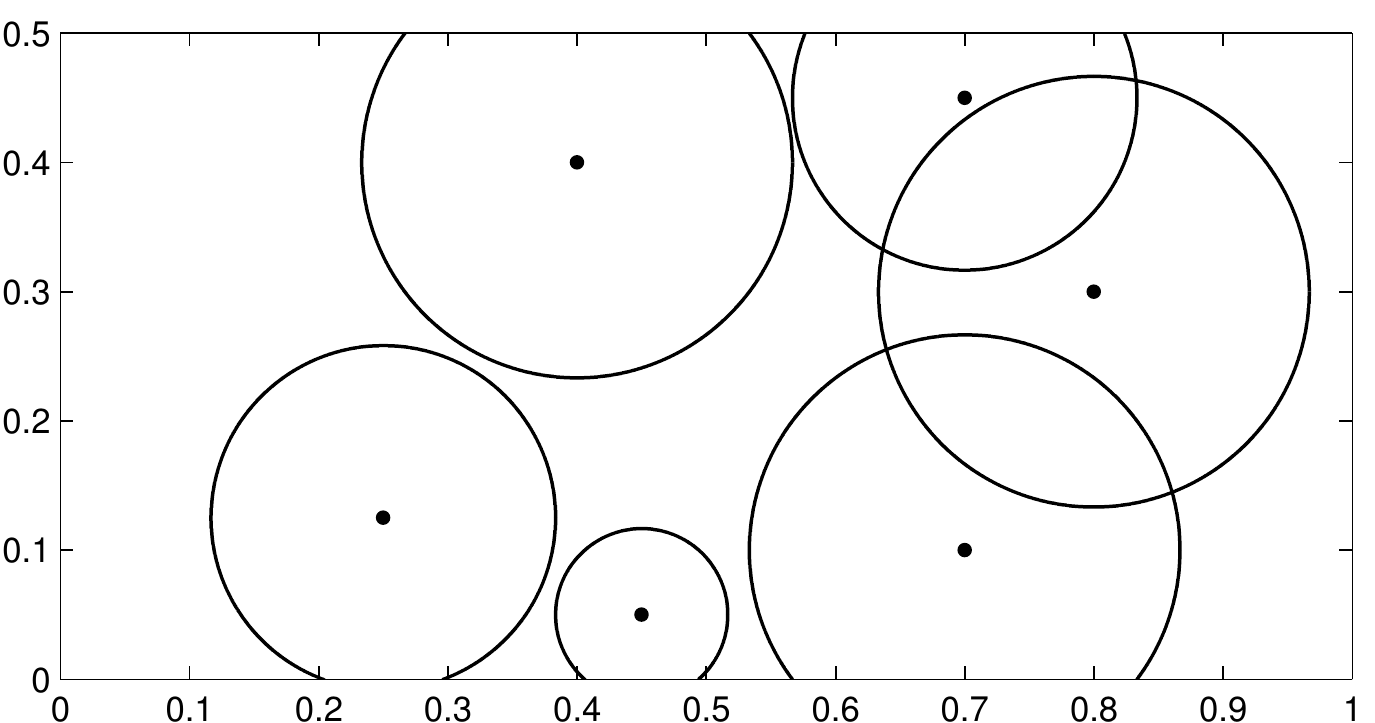}
\caption{Conceptual space divided into concepts according to a prototype-threshold approach. Some points in the space correspond to more than one concept, and some correspond to none.}
\label{fig:cplot}
\end{figure}

Within this framework, agents use sets of labels $LA = \{L_1, L_2, ..., L_n\}$ to describe an underlying conceptual space $\Omega$ with distance metric $d(x, y)$ between points. The conceptual space could be, as mentioned, the RGB colour space. Labels $L_i$ would then be concepts such as `red', `blue', `purple', `orange' and so on. These labels are viewed as regions of the conceptual space. So the concept `blue' is represented by the blue region in the colour cube. Within label semantics, these regions are specified by prototypes $P_i$ and thresholds $\e_i$. This is in contrast to G\"{a}rdenfors' original approach which is to view the space as partitioned by a Voronoi tessellation. If this latter approach is taken, each individual point in the conceptual space is allocated to exactly one label. With a prototype-threshold approach, it is easy to accommodate the idea of an object being accurately described by more than one concept, or conversely, some points within the space not being assigned to any concept. This difference is illustrated in figures \ref{fig:vplot} and \ref{fig:cplot}.

\begin{figure}
	 \centering
	  \def\svgwidth{0.3\textwidth}
	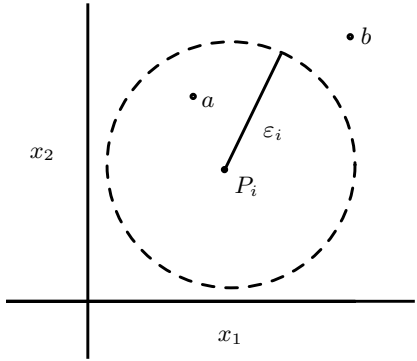

	  \caption{Prototype-threshold representation of a concept $L_i$. The conceptual space has dimensions $x_1$ and $x_2$. The concept has prototype $P_i$ and threshold $\e_i$. The uncertainty about the threshold is represented by the dotted line. The neighbourhood $\mathcal{N}^{\e_i}_{L_i}$ is the area within the dotted line. Element $a$ in the conceptual space is within the threshold, so it is appropriate to assert `$a$ is $L_i$'. Element $b$ is outside the threshold, so it is not appropriate to assert `$b$ is $L_i$'}
	\label{fig:pnth}
\end{figure}

 In this model, however, agents are uncertain as to exactly where the thresholds lie. To illustrate this, consider the concept `tall'. It is easy to point out a tall person, and to point out a person who is not tall, but it is difficult to specify the exact threshold between `tall' and `not tall'. This uncertainty concerning where the threshold lies is represented in the label semantics framework by saying that a threshold $\e_i$ is drawn from a probability distribution $\delta_i$. Labels $L_i$ are associated with neighbourhoods $\mathcal{N}^{\e_i}_{L_i} = \{\vec{x} \in \Omega : d(\vec{x}, P_i) \leq \e_i\}$, i.e. the region within the threshold. These ideas are represented in figure \ref{fig:pnth}.

The threshold $\e_i$ is uncertain, however, so there is some probability that $\e_i$ in figure \ref{fig:pnth} is actually wide enough to include the object $b$, i.e. that $L_i$ is appropriate to describe $b$. The appropriateness $\mu_{L_i}(x)$ of a label $L_i$ to describe an element $x$ is then given by the probability that $x$ lies within the neighbourhood $\mathcal{N}^{\e_i}_{L_i}$, i.e. that the distance $d(x, P_i)$ is less than $\e_i$. So:

\[
\mu_{L_i}(x) = P(d(x, P_i) \leq \e_i) = \int_{d(x, P_i)}^\infty \delta_i(\e_i)\mathrm{d}\e_i
\]

\begin{figure}[h]
\centering
\includegraphics[width = 0.35\textwidth]{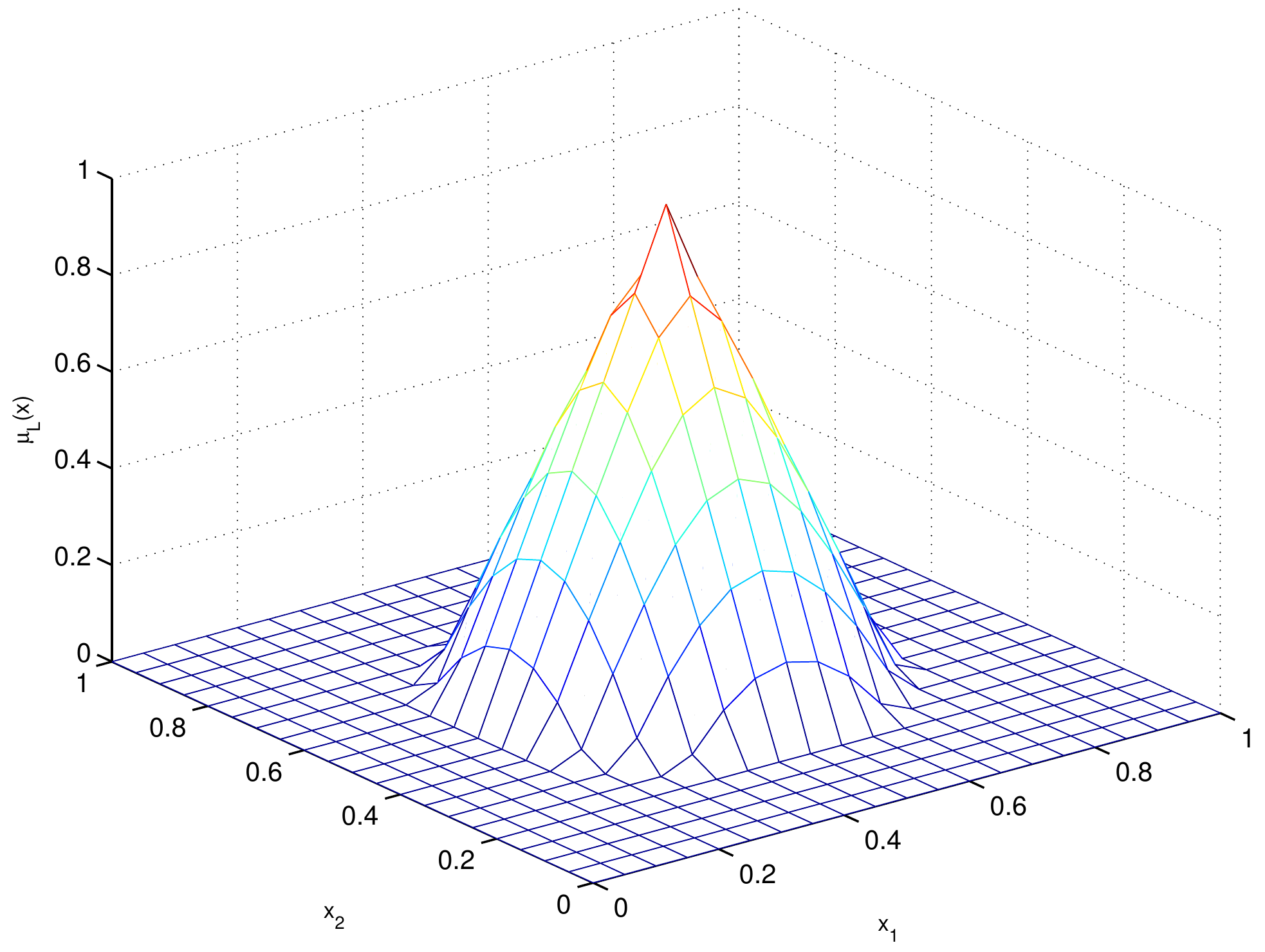}
\caption{Membership in a concept. The prototype of the label is at $[0.5, 0.5]$ and the threshold $\e$ has distribution $U[0,0.3]$. When $x = [0.5,0.5]$, $\mu_L(x) = 1$. As we move further away from the prototype, membership in the concept decreases, and is $0$ when $d(x, P_i) > 0.3$}
\label{fig:mu2d}
\end{figure}

Figure \ref{fig:mu2d} shows how this appropriateness measure works in a setup similar to that in figure \ref{fig:pnth}. 

This appropriateness measure is similar to Zadeh's description of fuzzy membership in a concept \cite{zadeh1965}.

\subsection{Linguistic hedges}
Hedges are words or phrases such as `very', `quite', `strictly speaking' which modify the domain of application of a concept. In particular, `very', and `quite' respectively contract or expand the domain of application of a concept,  so that, for example, `very tall' applies to fewer people than does `tall', whereas `quite tall' applies to more. Within fuzzy set theory, we expect that $\mu_{\text{very } L} (x) \leq \mu_L (x)$ and $\mu_{\text{quite } L} (x) \geq \mu_L (x)$. Applying this to the concept `tall', again,  this means that membership in the concept `very tall' should always be less than membership in `tall'. So anyone who can be described as `very tall' can also be described as `tall'.  Zadeh \cite{zadehhedges} uses operations of \emph{concentration} and \emph{dilation} to render these ideas. Concentration is described as $CON(\mu_{L_i}(x)) = (\mu_{L_i}(x))^2$ and dilation is often rendered as $DIL(\mu_{L_i}(x)) = (\mu_{L_i}(x))^{1/2}$. However, we argue, as do \cite{bosc}, that Zadeh's formulae are, to an extent, arbitrary, since the notion of taking a power of a membership value does not correspond to anything that language users might do. Rather, it simply has some of the right effects. As with \cite{bosc}, we take a semantic approach.

\begin{figure}
	 \centering
	  \def\svgwidth{0.3\textwidth}
	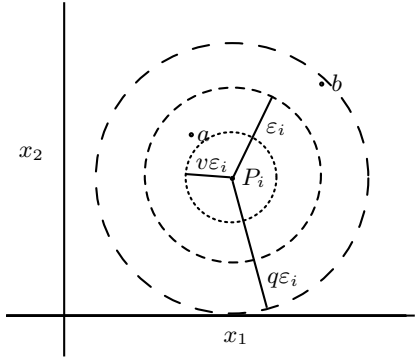
	  \caption{Representation of `very $L_i$' and `quite $L_i$'. `Very $L_i$' has prototype $P_i$ and threshold $v\e_i \leq \e_i$. `Quite $L_i$' has prototype $P_i$ and threshold $q\e_i \geq \e_i$. Notice that although $L_i$ is appropriate to describe $a$, $vL_i$ is not. Also, although $L_i$ is not appropriate to describe $b$, $qL_i$ is.}
	\label{fig:vq}
\end{figure}

In \cite{lewis2014}, we propose that a concept `very $L$' or `quite $L$' be rendered by considering that the prototype of `very/quite $L$' is equal to that of the base concept $L$, but that the threshold of the hedged concept `very/quite $L$' is respectively smaller or larger than that of the base concept. This approach is grounded in the idea that `very/quite $L$' should apply to respectively fewer or more objects than $L$. Narrowing or widening the threshold achieves this in a natural way. This is illustrated in figure \ref{fig:vq}. 

Our model of the hedges `very' and `quite' therefore requires simply that $v\e_i \leq \e_i$ and that $q\e_i \geq \e_i$. We implement this model in a version of the multi-agent simulation created in \cite{ettie} in order to investigate how the use of these hedges in a model of language helps the emergence of shared categories across a community of language users.

\section{METHODS}
\label{sec:method}
\subsection{Overview}
To investigate the utility of hedged assertions we implement a multi-agent simulation of a version of the category game \cite{belp}, following \cite{ettie}, in which shared categories develop over time as a result of the interactions of the category users. An overview of the game is as follows. Agents use labels to describe a conceptual space $\Omega$. At each timestep, agents are randomly paired into speakers and listeners, and each pair is shown a distinct element $x \in \Omega$. The speaker makes an assertion $\theta$ about the element based on its label set. The listener then updates its own label set to be more similar to that of the speaker, based on this assertion and a parameter $w$ which can be thought of as the age of the speaker. The update made by the speaker is a combination of shifting the prototype of the relevant label and changing the size of the threshold. The aim is that after a number of timesteps, label sets across the population have converged to a common set of shared categories.

\subsection{Conceptual models}
Each agent is equipped with the same number $n$ of labels $L_i$, with point prototypes $P_i \in \Omega$, where $\Omega  = [0,1]^3$. At the start of the simulations the $P_i$ are uniformly distributed around the space. Thresholds $\e_i$ are also randomly initiated, and considered to be some multiple of a base threshold $\e$. Each threshold $\e_i \sim U(0, b_i)$, where again, the $b_i$ can be considered to be a multiple of some common $b$, and the $b_i$ are taken from $U[0.5, 2]$. The distance metric is Euclidean. 

Each agent therefore has a label set $LA= \{L_1, L_2, ... L_n\}$. These labels can be hedged to form a set $LA^+ = LA \cup \{ \text{very }L_i, \text{quite }L_i : i = 1,..., n\}$. Hedged concepts have the same prototype $P_i$ as basic labels, but a scaled threshold $v\e_i$ or $q\e_i$ where  $v < 1$ and $q >1$. Agents can assert positive or negated, hedged or basic labels, giving an assertion set $AS = \{kL_i, \neg kL_i : i = 1, ..., n; k = \text{very}, \text{ quite}, \text{ basic}\}$, where $k = \text{basic}$ means that the label is not hedged.

\subsection{Assertion model}
At each timestep, half the agents are designated speaker agents and make assertions, determined by the assertion model used.The assertion model is based on the probability of making a particular assertion $\theta$, given that the object being described is $x$. Following methods in \cite{ettie, lawry2009}, we calculate the posterior probability of each $\theta \in AS$, given an element $x \in \Omega$. The assertion made by a speaker agent is the assertion with the highest probability. The posterior probability of each $\theta$, given $x$, is determined by the appropriateness of the assertion $\theta$ to describe $x$, i.e. $\mu_\theta(x)$, and the prior probability $P(\theta)$ of asserting $\theta$.

We first consider which sets of labels that are appropriate to describe $x \in \Omega$. The probability that any particular set of labels $F\subseteq LA$ are appropriate to describe $x$ is given by a probability mass function $m_x: 2^{LA} \rightarrow [0, 1]$. One way of determining $m_x$ is via the \emph{consonant selection function} introduced in \cite{lawry2009}. This states:

\begin{dfn}[Consonant selection function]
Given non-zero appropriateness measures on basic labels $\mu_{L_i}(x): i = 1,..., n$ ordered such that $\mu_{L_i}(x) \geq \mu_{L_{i+1}}$ for $i = 1, ..., n$, the consonant selection function identifies the mass function 
\begin{align*}
&m_x(\{L_1, ... , L_n\}) = \mu_{L_n}(x) \\
&m_x(\{L_1, ... , L_i\}) = \mu_{L_i}(x) - \mu_{L_{i+1}}(x) \text{ for } i = 1, .. n-1 \\ 
&m_x(\emptyset) = 1 - \mu_{L_1}(x)\\
&m_x(F) = 0 \text{ if }F \neq \{L_1, L_2, ..., L_k\} \text{ for some } k \leq n 
\end{align*}
\end{dfn}

Because we have ordered the labels by $\mu_{L_i}(x) \geq \mu_{L_{i+1}}$, if the label $L_i$ is appropriate to describe $x$, all labels $L_j : j < i$ must also be appropriate to describe $x$. The quantity $\mu_{L_i}(x) - \mu_{L_{i+1}}(x)$ corresponds to the idea that $x$ in some sense lies between the thresholds $\e_{i+1}$ and $\e_i$, so that $L_i$ is appropriate to describe $x$, but $L_{i+1}$ is not.  We extend this definition to the case of hedged labels simply by considering all hedged labels as basic labels, explained in the example below.

\begin{exa}[Determining the mass function]
\label{exa:mf}
Suppose we are determining the mass function for subsets $F \subseteq \{kL_1, kL_2: k = \text{very}, \text{ quite}, \text{ basic}\}$, given the point $a \in x_1 \times x_2$, as illustrated in figure \ref{fig:mf}.
\begin{figure}
	 \centering
	  \def\svgwidth{0.4\textwidth}
	 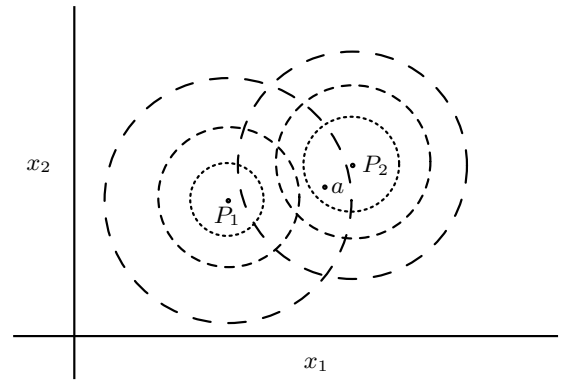
	  \caption{Determining the mass function on subsets of $\{kL_1, kL_2: k = \text{very}, \text{ quite}, \text{ basic}\}$. $P_1$ and $P_2$ represent prototypes for each lable $L_1$ and $L_2$, and the dotted lines give the different thresholds according to the hedges, as in figure \ref{fig:vq}. Notice that `quite $L_2$', $L_2$, `very $L_2$' and `quite $L_1$' are all appropriate to describe $a$, although with different appropriateness measures (not shown), but $L_1$ and `very $L_1$' are not.}
	\label{fig:mf}
\end{figure}

Suppose that $\mu_{\text{quite }L_2}(a) = 0.9$, $\mu_{L_2}(a) = 0.7$, $\mu_{\text{quite }L_1}(a) = 0.3$, $\mu_{\text{very }L_2}(a) = 0.1$, $\mu_{L_1}(a) = 0$, $\mu_{\text{very }L_1}(a) = 0$, giving us the order $\mu_{\text{quite }L_2}(a) \geq \mu_{L_2}(a) \geq \mu_{\text{quite }L_1}(a) \geq \mu_{\text{very }L_2}(a) \geq \mu_{L_1}(a) \geq \mu_{\text{very }L_1}(a)$. We may then assign probabilities to subsets of labels according to the consonant selection function:
\begin{align*}
m_x(F_6) &= m_x(\{\text{quite }L_2, L_2, \text{quite }L_1, \text{very }L_2, L_1, \text{very }L_1\}) \\
&= \mu_{\text{very }L_1}(a) = 0\\
m_x(F_5) &= m_x(\{\text{quite }L_2, L_2, \text{quite }L_1, \text{very }L_2, L_1\}) \\
&= \mu{L_1}(a) -\mu_{\text{very }L_1}(a) = 0\\
m_x(F_4) &= m_x(\{\text{quite }L_2, L_2, \text{quite }L_1, \text{very }L_2\}) \\
&= \mu_{\text{very }L_2}(a) - \mu_{L_1}(a) = 0.1\\
m_x(F_3)&= m_x(\{\text{quite }L_2, L_2, \text{quite }L_1\}) \\
&= \mu_{\text{quite }L_1}(a) - \mu_{\text{very }L_2}(a) = 0.2\\
m_x(F_2) &= m_x(\{\text{quite }L_2, L_2\}) = \mu_{L_2}(a) - \mu_{\text{quite }L_1}(a) = 0.4\\
m_x(F_1) &= m_x(\{\text{quite }L_2\}) = \mu_{\text{quite }L_2}(a) - \mu_{L_2}(a) = 0.2\\
m_x(\emptyset) &= 1 - \mu_{\text{quite }L_2}(a) = 0.1
\end{align*}
\end{exa}

Having determined the probability mass function on sets of labels, a mass assignment on sets of assertions is then defined.

\begin{dfn}[Mass assignment on assertions]

$ma_x: 2^{AS} \rightarrow [0, 1]$ is defined such that:
\[
ma_x(G) = \sum_{F \subseteq LA^+ : \mathscr{C}(F) = G} m_x(F)
\]
where  $\mathscr{C}(F) = \{\theta \in AS : F\in \lambda(\theta)\}$, and $\lambda(\theta)$ is defined recursively by 

 \begin{align*}
\lambda(kL_i) &= \{F \subseteq LA^+: kLi \in F\}\\
\lambda(\neg \theta) &= (\lambda(\theta))^c\\
\lambda(\theta \wedge \phi) &= \lambda(\theta) \cap \lambda(\phi)\\
\lambda(\theta \vee \phi) &= \lambda(\theta) \cup \lambda(\phi)\\
\end{align*}
\end{dfn}

This definition has the implication that for $G_i = F_i \cup \{\neg kL_j: kL_j \in LA^+\backslash F_i\}$, $ma_x(G_i) = m_x(F_i)$.

Then the probability of an assertion $\theta$ being made, given that an object $x$ is being described, can be calculated by summing over $G \subseteq AS$ that contain $\theta$.

\begin{dfn}
Given a prior distribution on $AS$, a posterior distribution given an object $x$ can be calculated by:
 \begin{align*}
P(\mathscr{A} = \theta | x) &= \sum_{G \subseteq AS: \theta \in G} ma_x(G)P(\mathscr{A} = \theta |\mathscr{A} \in G) \\
&= P(\theta) \sum_{G \subseteq AS: \theta \in G} \frac{ma_x(G)}{P(G)}
 \end{align*}
\end{dfn}
Here, $P(G) = \sum_{\varphi \in G} P(\varphi)$

The value of $P(\theta)$ for one particular label $L$ is a product of two elements: the prior probability $pp$ of making a positive assertion (or $1 - pp$ for a negated assertion); and the prior probability of making a hedged assertion, given by $pv$ for making an assertion hedged with the word `very', $pq$ for making an assertion hedged with `quite' or $1 - pv - pq$ for making a basic assertion, summarised in table \ref{tab:ptab}.

\begin{table}
	\centering
	\caption{Prior probabilities of each type of assertion $\pm kLi$}
	\begin{tabular}{cccc}
	\hline
	$*$ & $pv$ & $pb$ & $pq$\\ \hline
	$pp$ & $P(vL)$ & $P(L)$ & $P(qL)$\\ \hline
	$pn$ & $P(\neg vL)$ & $P(\neg L)$ & $ P(\neg qL)$ \\ \hline
	\end{tabular}
	\label{tab:ptab}
\end{table}

The prior probability of asserting any particular label $L_i \in LA$ is uniform across $LA$. Hence the value of $P(\theta)$ calculated above should be divided by $n$, giving, for example, 
\[
P(\neg vL_2) = \frac{pn*pv}{n}
\]

\begin{exa}[Determining the posterior probability of assertion]
Suppose, for an easy example, we want to calculate the probability of asserting `very $L_1$', given object $a$, as in example \ref{exa:mf}. We need to calculate 

\[
P(\mathscr{A} = \text{`very $L_1$'} | a) = P(\theta) \sum_{G \subseteq AS: \text{`very $L_1$'} \in G} \frac{ma_x(G)}{P(G)}
\]

where $G_i = F_i \cup \{\neg kL_j: kL_j \in LA^+\backslash F_i\}$. However, the only subset $G_i \ni$ `very $L_1$' is $G_6$, so 

\begin{align*}
P(\mathscr{A} = \text{`very $L_1$'} | x) &= P(\text{`very $L_1$'}) \frac{ma_x(G_6)}{P(G_6)}\\
&=0
\end{align*}

Suppose, for a more involved example, the label set $LA^+$ is as in example \ref{exa:mf}, with $pp = 0.7$, $pv = 0.7$, $pq = 0.2$, and we want to determine $P(\mathscr{A} = \text{quite }L_1 | a)$. The prior probability $P(\text{quite }L_1) = \frac{0.7*0.2}{2} = 0.07$. So we have:

\begin{align*}
P(\mathscr{A} &= \text{quite }L_1 | a)\\
&= 0.07\sum_{G_i: \text{quite }L_1 \in G_i} \frac{ma_x(G_i)} {P(G_i)}\\
&= 0.07(\frac{ma_x(G_6)} {P(G_6)} + \frac{ma_x(G_5)} {P(G_5)}+\frac{ma_x(G_4)} {P(G_4)}+ \frac{ma_x(G_3)} {P(G_3)})\\
 &= 0.07(0 + 0 + \frac{0.1}{\sum_{\varphi \in G_4}P(\varphi)} + \frac{0.2}{\sum_{\varphi \in G_3}P(\varphi)})\\
 &= 0.07(\frac{0.1}{0.54} + \frac{0.2}{0.4})\\
 &= 0.048\\
\end{align*}
\end{exa}

Having calculated the probability of each assertion, the speaker agent makes the most probable assertion $\theta \in AS$.

\subsection{Updating algorithms}
Once the speaker agent has made assertion $\theta$, the listener agent computes $\mu_\theta(x)$ based on its current label set. If $\mu_\theta(x) < w$, where $w$ is a parameter that can be thought of as the age of the speaker agent, the listener agent updates its label set $LA$ by moving the prototype and/or changing the threshold of the concept, until $\mu_{\theta}(x) = w$. Formulae  for these updates are again based on \cite{ettie}. A label defined by $P_i$ and $\e_i$ is updated to $P_i' = P_i - \lambda(x - P_i)$ and $\e_i' = \alpha \e_i$. Values for $\lambda$ and $\alpha$ are sought, such that $\mu_{\theta}'(x) = w$.

\subsubsection{Case 1: $\theta = kL_i$}
Recall that $\e_i \sim U(0, b_i)$, so that for $x \in\Omega$,
\[
\mu_{kL_i}(x) = 1 - \frac{||x-P_i||}{kb_i} < w \text{ by assumption.} 
\]

The label $L_i$ is updated to $L_i'$, where $P_i' = P_i - \lambda(x - P_i)$ and $\e_i' = \alpha \e_i$, such that  $\mu_{kL_i'}(x) \geq w$, and minimising the distance between the interpretations as measured by the Haussdorff distance between the two neighbourhoods,
\begin{align}
\label{eq:hmet}
\mathscr{H}(\mathcal{N}_{L_i}, \mathcal{N}_{L_i'}) &= ||P_i - P_i'|| + |\e_i -\e_i'|\\
&= |\lambda|||x-P_i|| + \frac{\e b_i}{b}|1-\alpha| \text{ (*)} \nonumber
\end{align}

To minimise the update, we set $\mu_{kL_i'}(x) = w$, so:

\begin{align*}
w = \mu_{kL_i'}(x) = 1 - \frac{||x-P_i'||}{kb_i'} = 1 - \frac{|1-\lambda|||x-P_i||}{\alpha kb_i}
\end{align*}

which gives
\begin{align*}
\alpha &= \frac{|1-\lambda|||x-P_i||}{(1-w)kb_i}\\
&= \frac{(1-\lambda)||x-P_i||}{(1-w)kb_i} \text{\quad since $\lambda = 1 \rightarrow P_i' =x$}
\end{align*}

To update $L_i$ we will always want $\lambda \geq 0$, $\alpha \geq 1$, as we are dealing with a positive label.

Substituting $\alpha$ into equation (*),  we obtain
\begin{align}
\mathscr{H}&(\mathcal{N}_{L_i}, \mathcal{N}_{L_i}') = |\lambda|||x-P_i|| + \frac{\e b_i}{b}(\frac{(1-|\lambda|)||x-P_i||}{(1-w)kb_i} - 1)\nonumber \\
\label{eq:tominimise}
&= |\lambda|||x -P_i||(1 - \frac{\e}{b(1-w)k}) + \frac{\e ||x-P_i||}{b(1-w)k} - \frac{\e b_i}{b} 
\end{align}

Then if $1 - \frac{\e }{b(1-w)k} >0$, i.e. $\e < b(1-w)k$, the quantity (\ref{eq:tominimise}) can be minimised by setting $\lambda = 0$ so $\alpha = \frac{||x-P_i||}{(1-w)kb_i}$. Otherwise, we have $\alpha = 1$, $\lambda = 1 - \frac{(1-w)kb_i}{||x - P_i||}$. 

Since $\e$ is a random variable, so is the choice between $\lambda$ and $\alpha$. We therefore need a concrete updating rule. We update $P_i$ and $b_i$ with the expected values of $\lambda$ and $\alpha$ respectively. $\e \sim \text{Uniform}[0, b]$, so

\begin{align*}
P(\e < b(1-w)k) & = 
\begin{cases} 
(1-w)k & \text{if }  (1 - w)k < 1\\
1   & \text{otherwise\ }
  \end{cases}
\end{align*}

We can therefore calculate 
\begin{align*}
E(\alpha) & = 
\begin{cases} 
\frac{||x - P_i||}{b_i} + 1 - (1-w)k & \text{if }  (1 - w)k < 1\\
\frac{||x - P_i||}{(1-w)kb_i}   & \text{otherwise\ }
  \end{cases}
\end{align*}

and
\begin{align*}
E(\lambda) & = 
\begin{cases} 
(1 - (1-w)k )(1 -  \frac{(1-w)kb_i}{||x - P_i||} ) & \text{if }  (1 - w)k < 1\\
0   & \text{otherwise\ }
  \end{cases}
\end{align*}

\subsubsection{Case 2: $\theta = \neg kL_i$}
By an entirely similar argument, we obtain
\begin{align*}
E(\alpha) & = 
\begin{cases} 
\frac{||x - P_i||}{b_i} + 1 - wk & \text{if }  wk < 1\\
\frac{||x - P_i||}{wkb_i}   & \text{otherwise\ }
  \end{cases}
\end{align*}

and
\begin{align*}
E(\lambda) & = 
\begin{cases} 
(1 - wk )(1 -  \frac{wkb_i}{||x - P_i||} ) & \text{if }  wq < 1\\
0   & \text{otherwise\ }
  \end{cases}
\end{align*} 

So at each timestep, each listener agent, for whom $\mu_\theta(x) < w$, updates the relevant label using the the quantities $E(\alpha)$, $E(\lambda)$.

\subsection{Performance metrics}
Performance metrics from \cite{ettie} are used, measuring the Average Pairwise Distance between label sets (APD) and the Average Label Overlap (ALO).  APD measures the difference in label sets in the community, and ALO indicates the extent to which an agent's concepts overlap. We seek low values for each metric.

APD is calculated using the Haussdorff distance between two neighbourhoods as given in equation \ref{eq:hmet}. The difference between the label sets of any one pair of agents is given by

\[
IPD = \sum_{i = 1}^n\mathscr{H}(\mathcal{N}_{L_i}^j, \mathcal{N}_{L_i}^k)
\]

where $n$ is the number of labels each agent has and $j$ and $k$ refer to distinct agents.

This is averaged over pairs of agents. There are $N$ agents, therefore $\binom{N}{2}$ pairs, giving:

\[
APD = \frac{2}{N(N-1)}\sum_{k = j + 1}^N\sum_{j = 1}^N IPD_{jk}
\]

ALO is the extent to which labels overlap. To calculate this, we take the maximum value of the intersection of a pair of labels, as measured by a min rule. We average this value over pairs of labels. The overlap within an individual's label set is therefore

\[ 
ILO = \frac{2}{n(n-1)}\sum_{j = i + 1} ^ n \sum_{i = 1}^n max\{min\{\mu_{L_i}(x),\mu_{L_j}(x) : x \in \Omega \}\}
\]

Averaged across the population this is:
\[
ALO = \frac{1}{N}\sum_{k = 1}^N ILO_k
\]
where $ILO_k$ siginifies agent $k$'s label overlap.

\subsection{Simulation process}
\label{sec:simproc}
Simulations with $n = 100$ agents were run for $T = 10^4$ timesteps. Agent weights $w \in [0.2, 0.8]$ were updated at each timestep in increments of $1/T$. When $w \geq 0.8$, agents are reborn with randomised labels and $w = 0.2$. 20 simulations are run for each reported combination of parameters.

\cite{ettie} show that if $pp \in [0.5, 0.6]$ then performance of the system changes from low ALO and high APD to vice versa at approximately $pp = 0.56$. We ran simulations in a slightly extended range for comparison, varying the prior probabilities $pv$, $pb$ and $pq$ of asserting the different hedges `very', `basic', and `quite'. We present results from three sets of parameters. As a baseline we run simulations with no hedges, i.e. $pv = 0$, $pb = 1$, $pq = 0$. To investigate the effects of using contraction hedges, we run simulations with parameters $pv = 0.7$, $pb = 0.2$, $pq = 0.1$. For expansion hedges, we use parameters $pv = 0.1$, $pb = 0.2$, $pq = 0.7$.

\section{RESULTS}
\label{sec:results}
The results presented show performance against the two metrics after $10^4$ simulation timesteps. By this point, the population has generally reached a steady state in which performance does not greatly change.  

\begin{figure}[h]
\centering
\includegraphics[width = 0.35\textwidth]{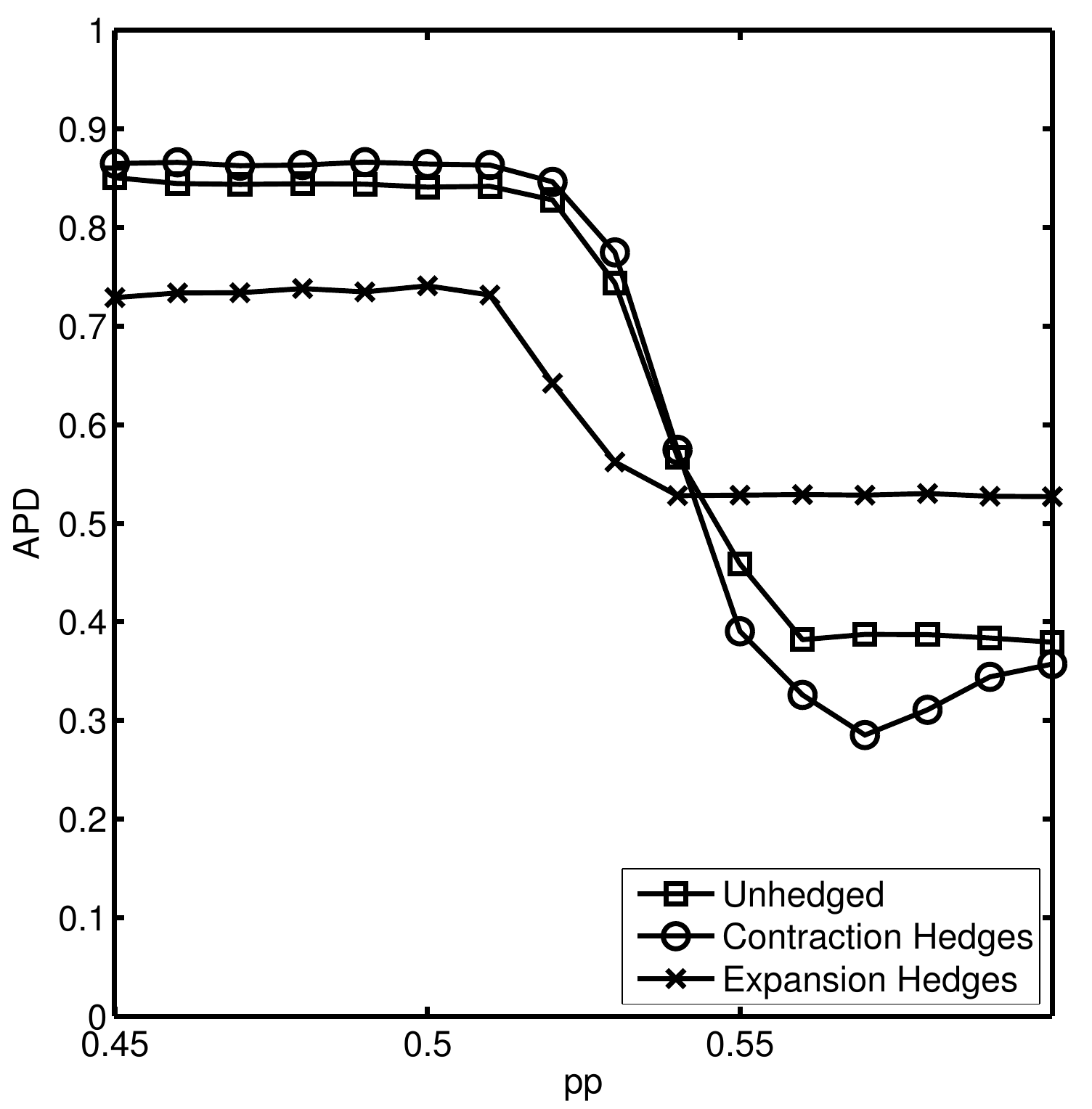}
\caption{APD after $10^4$ timesteps. Using contraction hedges reduces the minimum APD achieved from $0.38$ at $pp = 0.56$ to $0.29$ at $pp = 0.57$. Expansion hedges reduce APD from $0.85$ to $0.73$ at $pp =0.45$}
\label{fig:cmpAPDI}
\end{figure}

Figure \ref{fig:cmpAPDI} shows the steady state of APD achieved after $10^4$ timesteps for a range of values $pp \in [0.4, 0.6]$. Three sets of results are presented: results using unhedged assertions; results with a high prior probability of using contraction hedges; and results from simulations with a high prior probability of asserting expansion hedges, where these prior probabilities are as stated in \ref{sec:simproc}.

\begin{figure}[h]
\centering
\centering
\includegraphics[width = 0.35\textwidth]{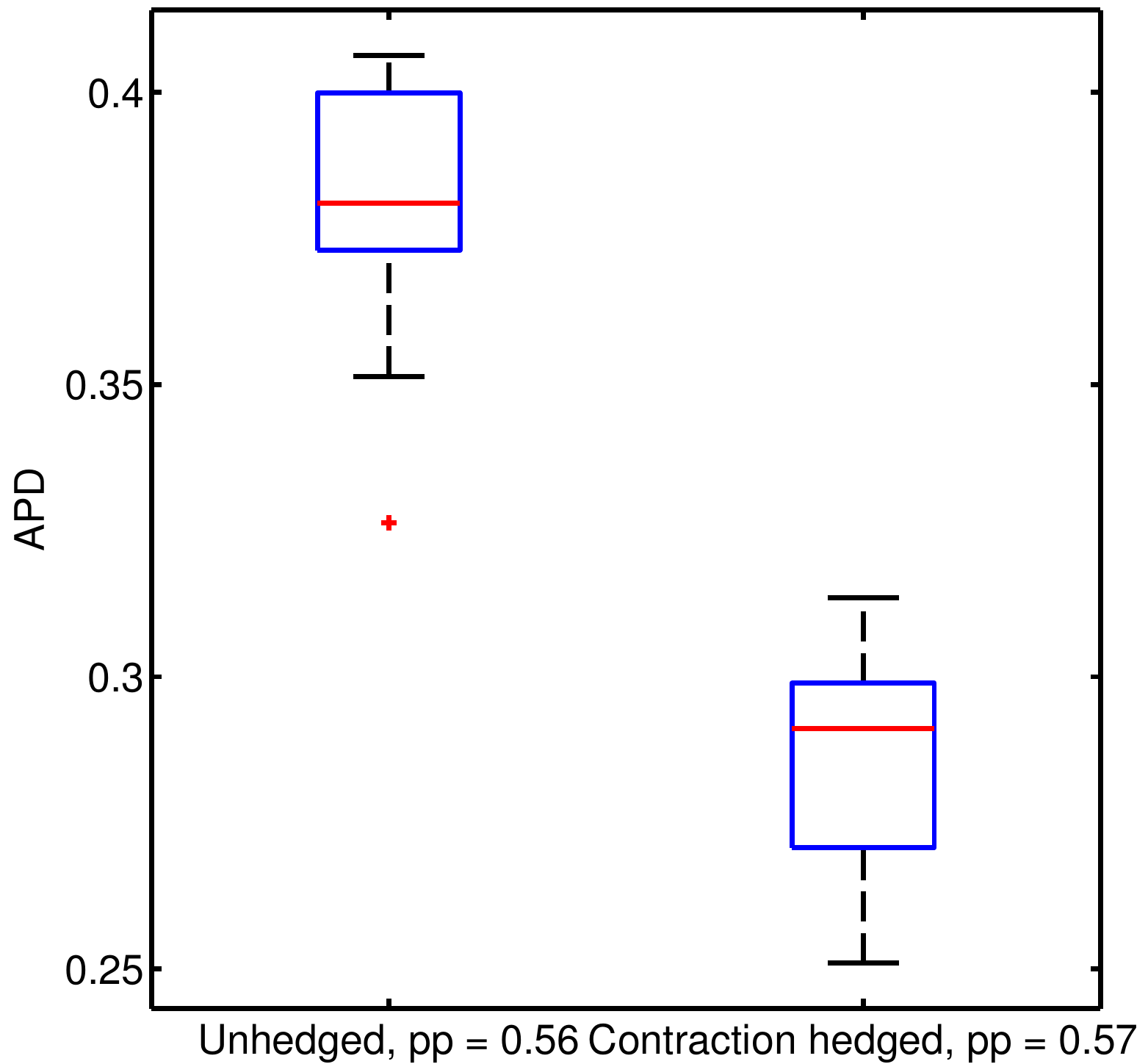}
\caption{Using contraction hedges reduces minimum APD. Box and whisker plot of APD after $10^4$ timesteps for 20 simulations, for $pp = 0.56$ unhedged, $pp = 0.57$ with a high probability of contraction hedges (values of $pp$ at which minimum APD is achieved). The middle line shows median value, the box shows the 25th and 75th percentile. Whiskers show the range of data excluding outliers, and crosses show outliers.}
\label{fig:chAPDI}
\end{figure}

\begin{figure}[h]
\centering
\includegraphics[width = 0.35\textwidth]{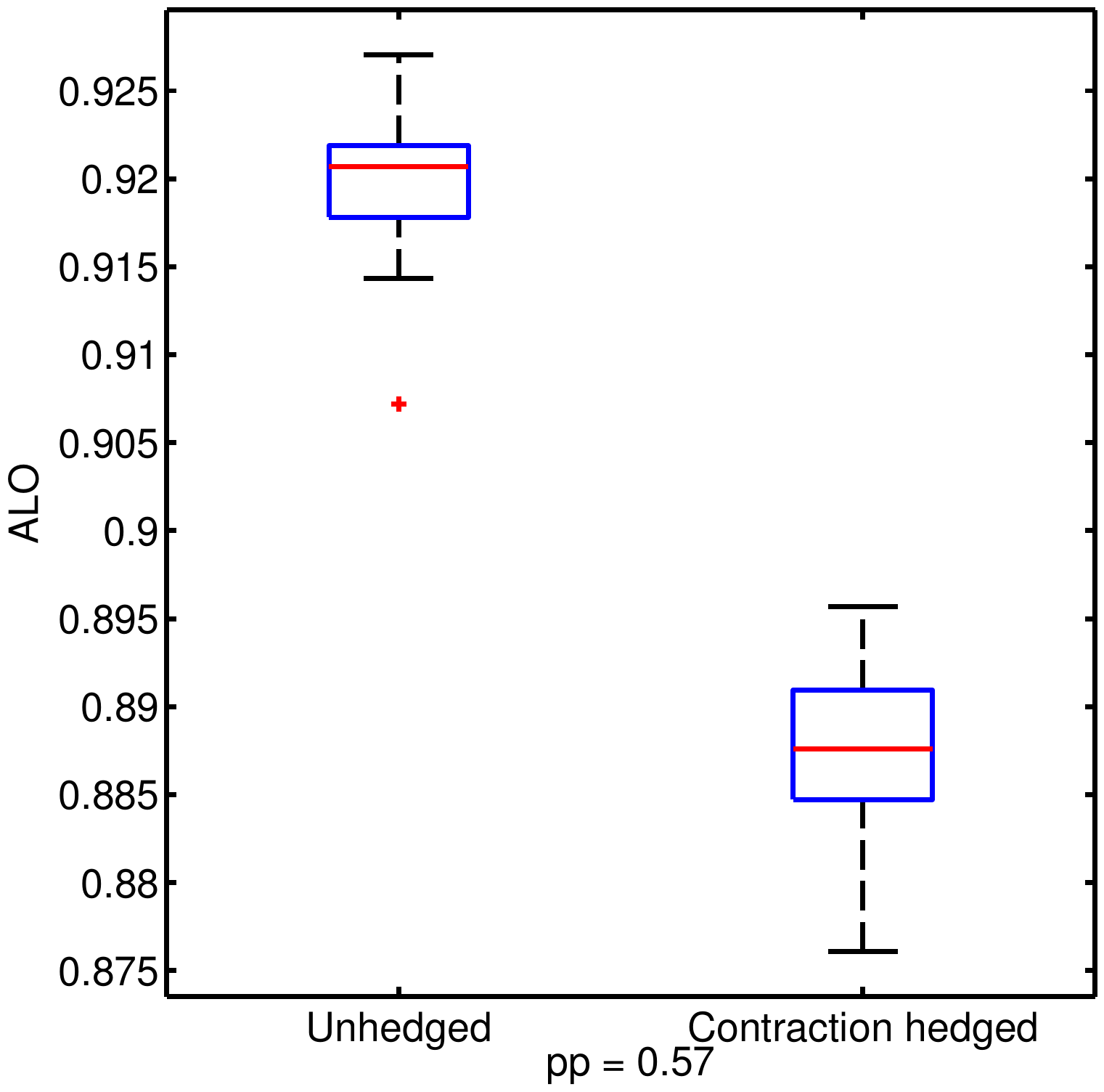}
\caption{Using contraction hedges slightly reduces ALO. Box and whisker plot of ALO after $10^4$ timesteps for 20 simulations, for $pp = 0.57$. The middle line shows median value, the box shows the 25th and 75th percentile. Whiskers show the range of data excluding outliers, and crosses show outliers.}
\label{fig:chALOP}
\end{figure}

\begin{figure}[h]
\centering
\includegraphics[width = 0.35\textwidth]{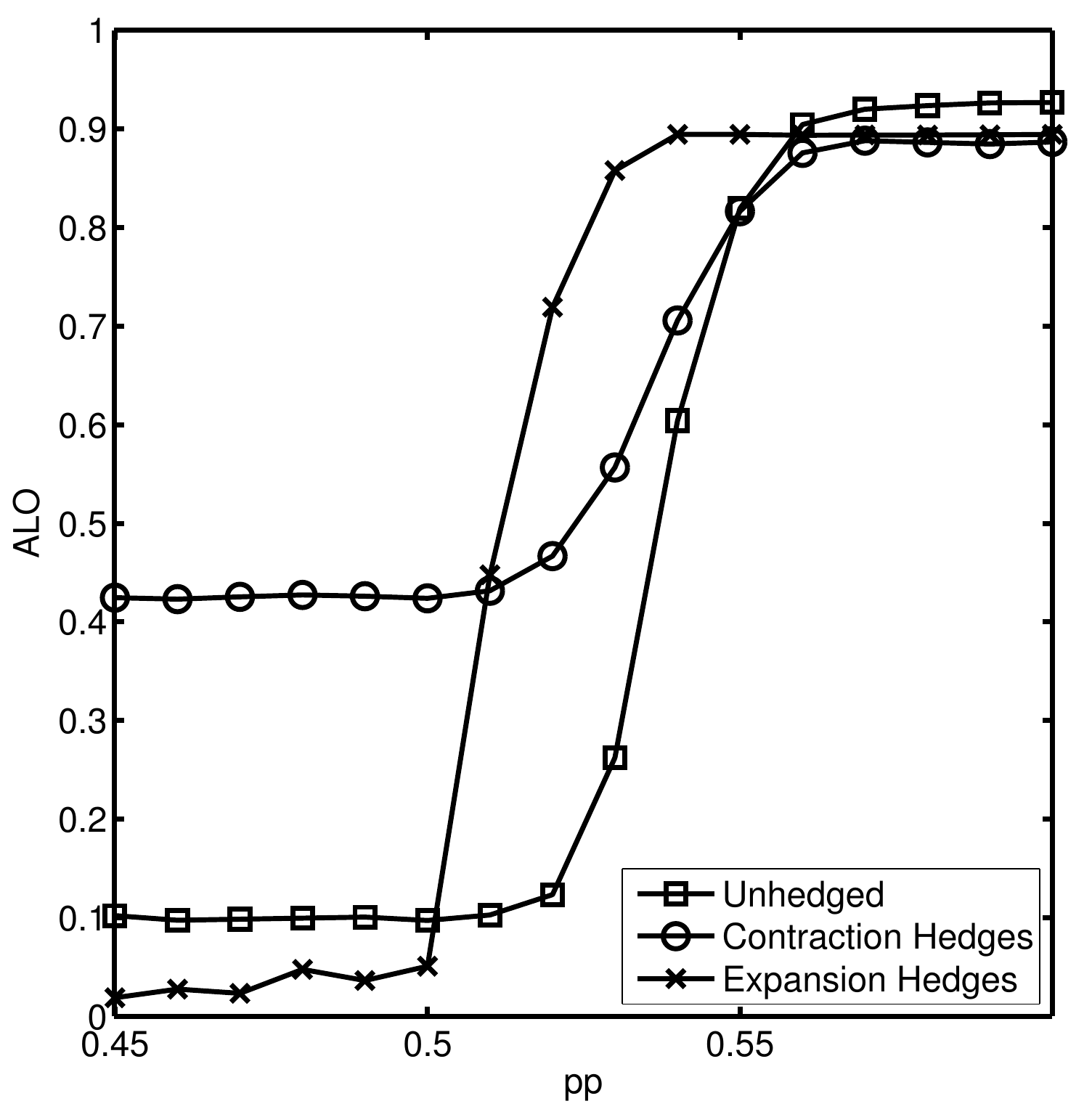}
\caption{Contraction hedges slightly reduce high levels of ALO. At $pp = 0.57$, ALO is reduced from $0.92$ to $0.89$. Expansion hedges reduce minimum ALO  from $0.1$ for unhedged assertions to $0.02$ for expansion hedged assertions, at $pp = 0.45$.}
\label{fig:cmpALOP}
\end{figure}

A high prior of asserting contracted labels reduces minimum APD achieved from $0.38$ when $pp=0.56$ or $pp = 0.6$ to  $0.29$ when $pp=0.57$ (figure \ref{fig:cmpAPDI}). Performing a paired t-test across the 20 simulations gives the mean difference between these values as $0.097$. This difference is statistically significant with $p < 0.001$ and with 95\% confidence interval $[0.084, 0.110]$. The median and range of results are given in figure \ref{fig:chAPDI}. At $pp=0.57$, ALO decreases, from $0.92$ to $0.89$ (figure \ref{fig:cmpALOP}). The mean value of this difference across the 20 simulations is $0.032$. Again, this is statistically significant with  $p < 0.001$ and 95\% confidence interval of $[0.029, 0.035]$, further illustrated in figure \ref{fig:chALOP}. These results imply that a high prior probability of asserting contraction hedges enables us to improve convergence between agents' label sets as well as reducing overlap within label sets slightly.

\begin{figure}[h]
\centering
\centering
\includegraphics[width = 0.35\textwidth]{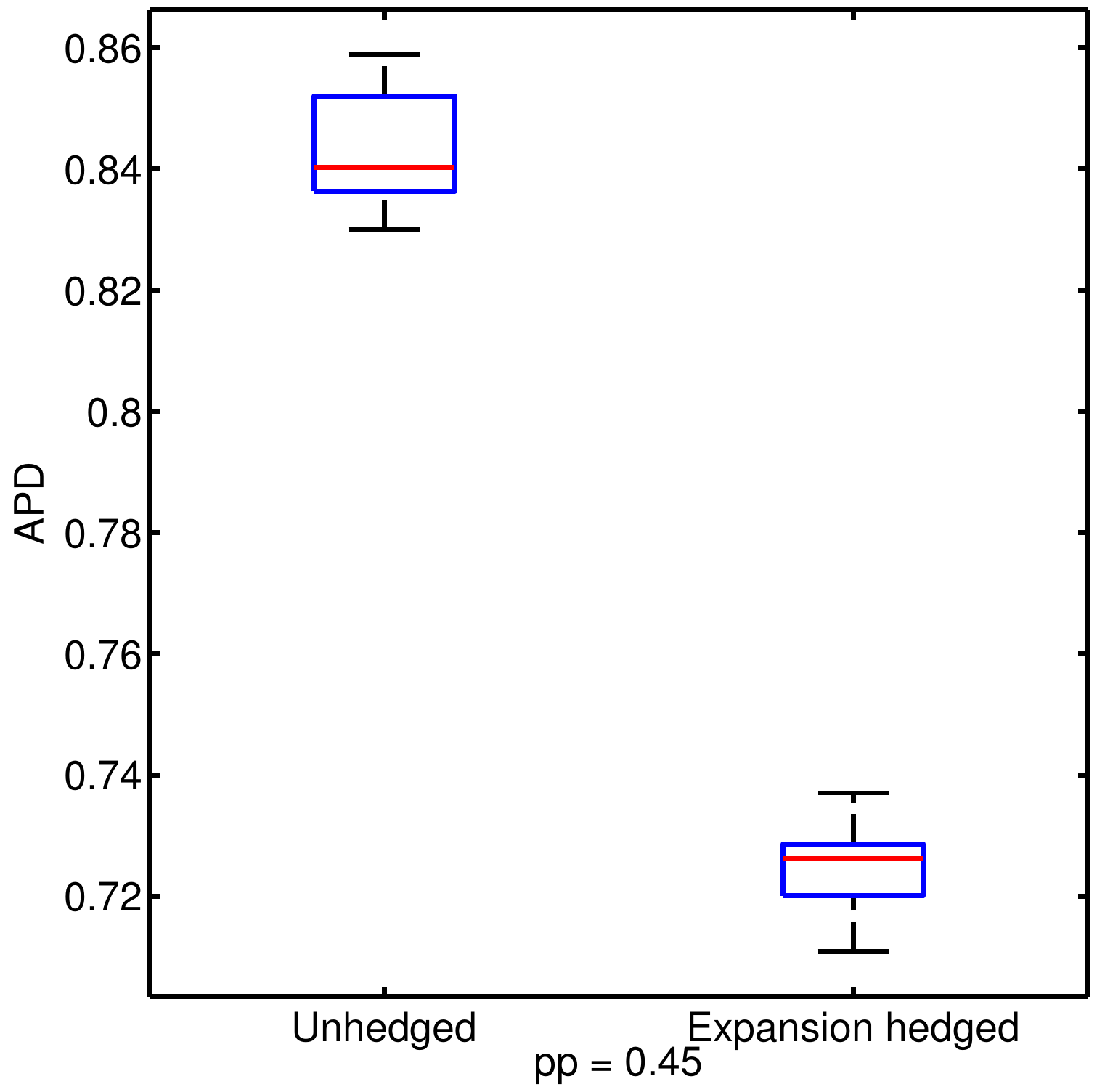}
\caption{Expansion hedges reduce maximum APD. Box and whisker plot of APD after $10^4$ timesteps for 20 simulations, for $pp = 0.45$.  The middle line shows median value, the box shows the 25th and 75th percentile. Whiskers show the range of data excluding outliers, and crosses show outliers.}
\label{fig:ehAPDI}
\end{figure}

\begin{figure}[h]
\centering
\includegraphics[width = 0.35\textwidth]{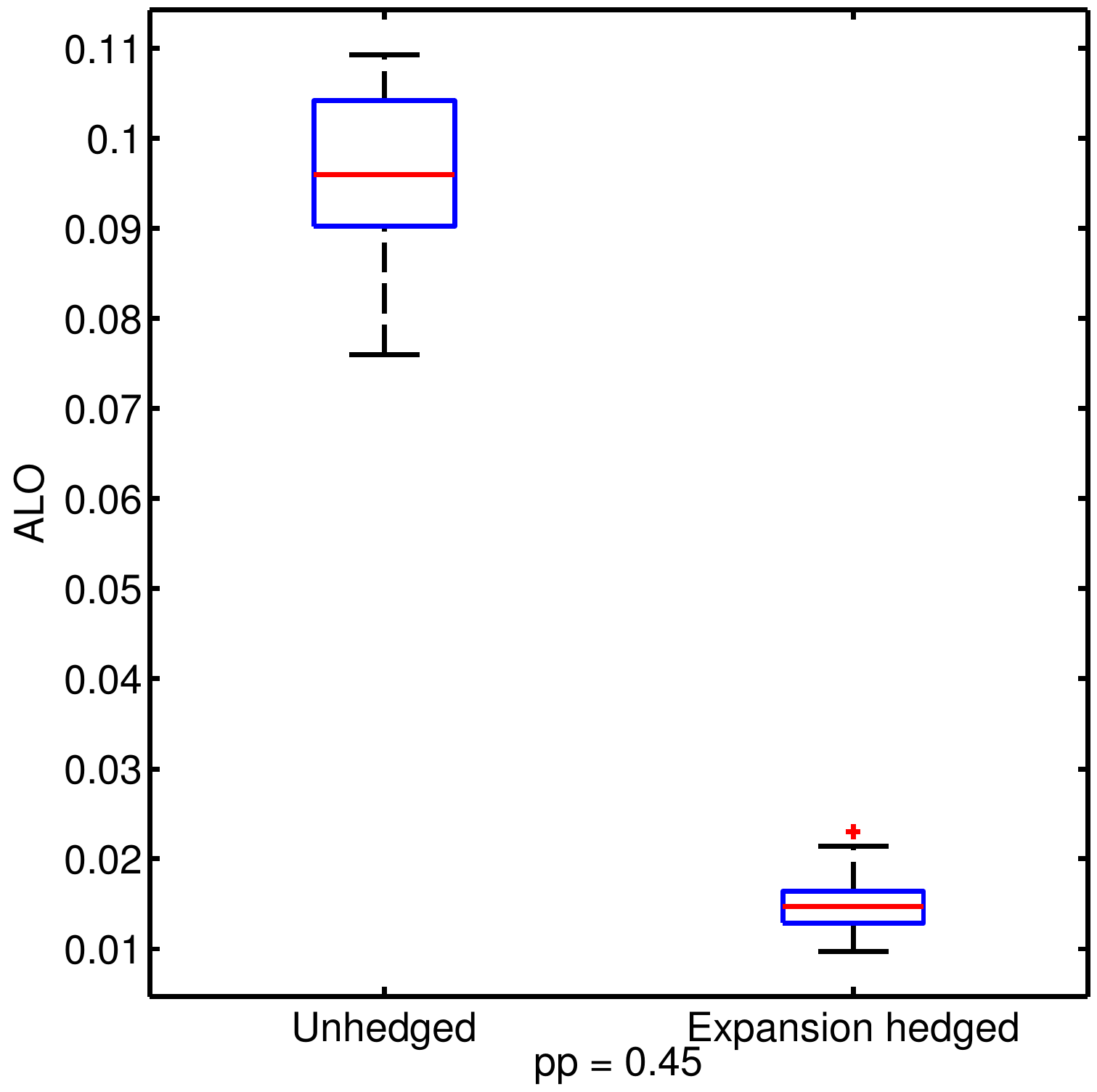}
\caption{Expansion hedges reduce minimum ALO. Box and whisker plot of ALO after $10^4$ timesteps for 20 simulations, for $pp = 0.45$. The middle line shows median value, the box shows the 25th and 75th percentile. Whiskers show the range of data excluding outliers, and crosses show outliers.}
\label{fig:ehALOP}
\end{figure}

With a high prior probability of asserting expanded labels, lower values of ALO can be achieved when the probability of asserting positive labels is $0.45$ , decreasing to $0.02$ compared to $0.1$, figure \ref{fig:cmpALOP}. The mean difference between these values across the 20 simulations is 0.083, which is statistically significant with $p < 0.001$ and a 95\% confidence interval of $[0.078, 0.089]$. The data is represented in figure \ref{fig:ehALOP}. At this value of $pp$, APD achieved is $0.73$ compared to $0.85$ for unhedged assertions, figure \ref{fig:cmpAPDI}. The mean value of this difference across the 20 simulations is $0.12$. This figure is statistically significant with $p < 0.001$ and 95\% confidence interval $[0.116, 0.126]$. The data is again represented in figure \ref{fig:ehAPDI}.  A high prior probability of asserting expansion hedges therefore enables minimal overlap to be maintained at low $pp$ whilst improving convergence. 

\begin{figure}[h]
\centering
\includegraphics[width = 0.35\textwidth]{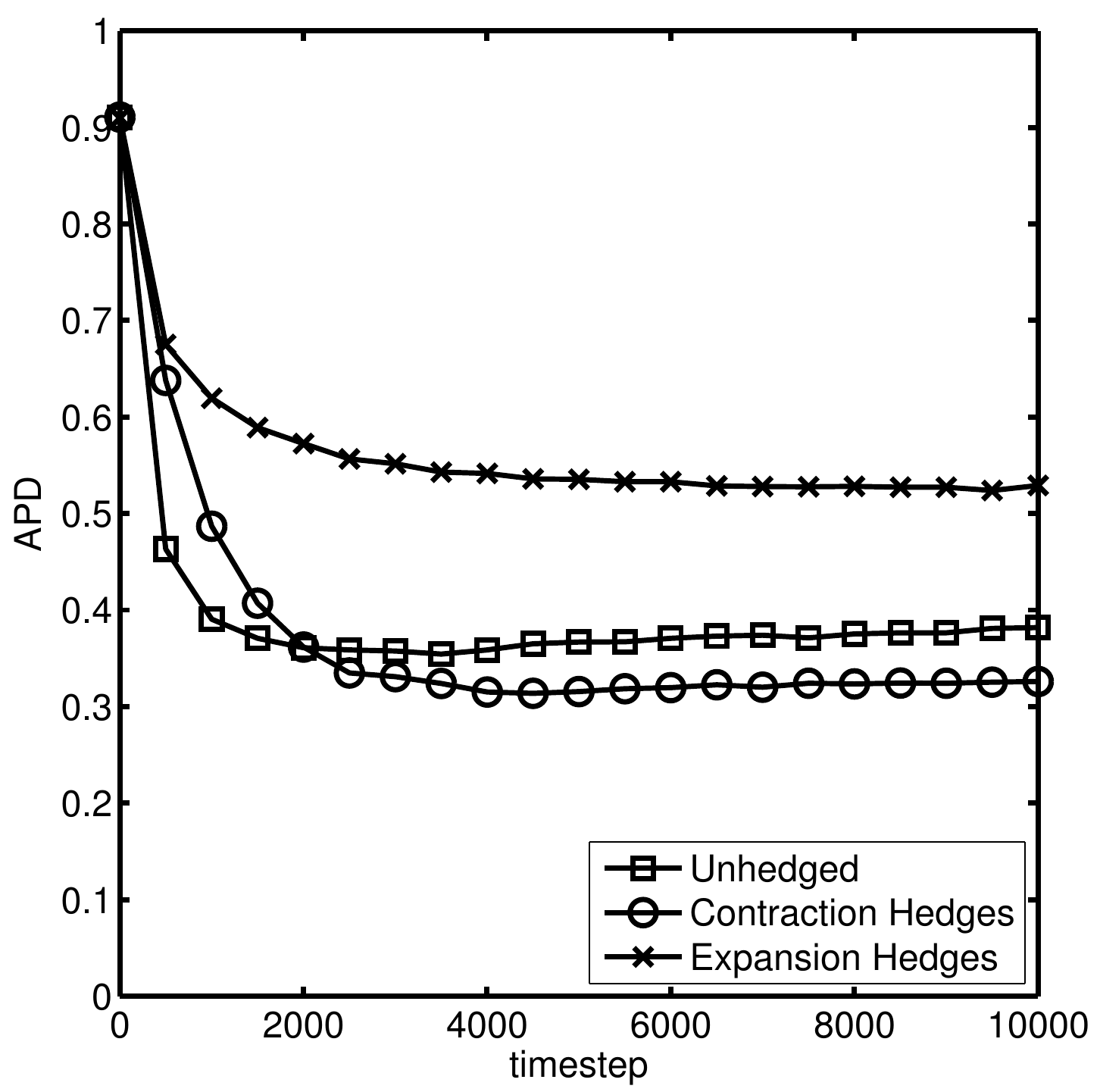}
\caption{APD vs time for models with no hedged assertions, contracted assertions and expanded assertions, for a prior probability of positive assertions $pp = 0.56$. Although the final value reached is lower when there is a high probability of making contracted assertions, the community of agents takes longer to reach that value. }
\label{fig:APDI56}
\end{figure}

\begin{figure}[h]
\centering
\includegraphics[width = 0.35\textwidth]{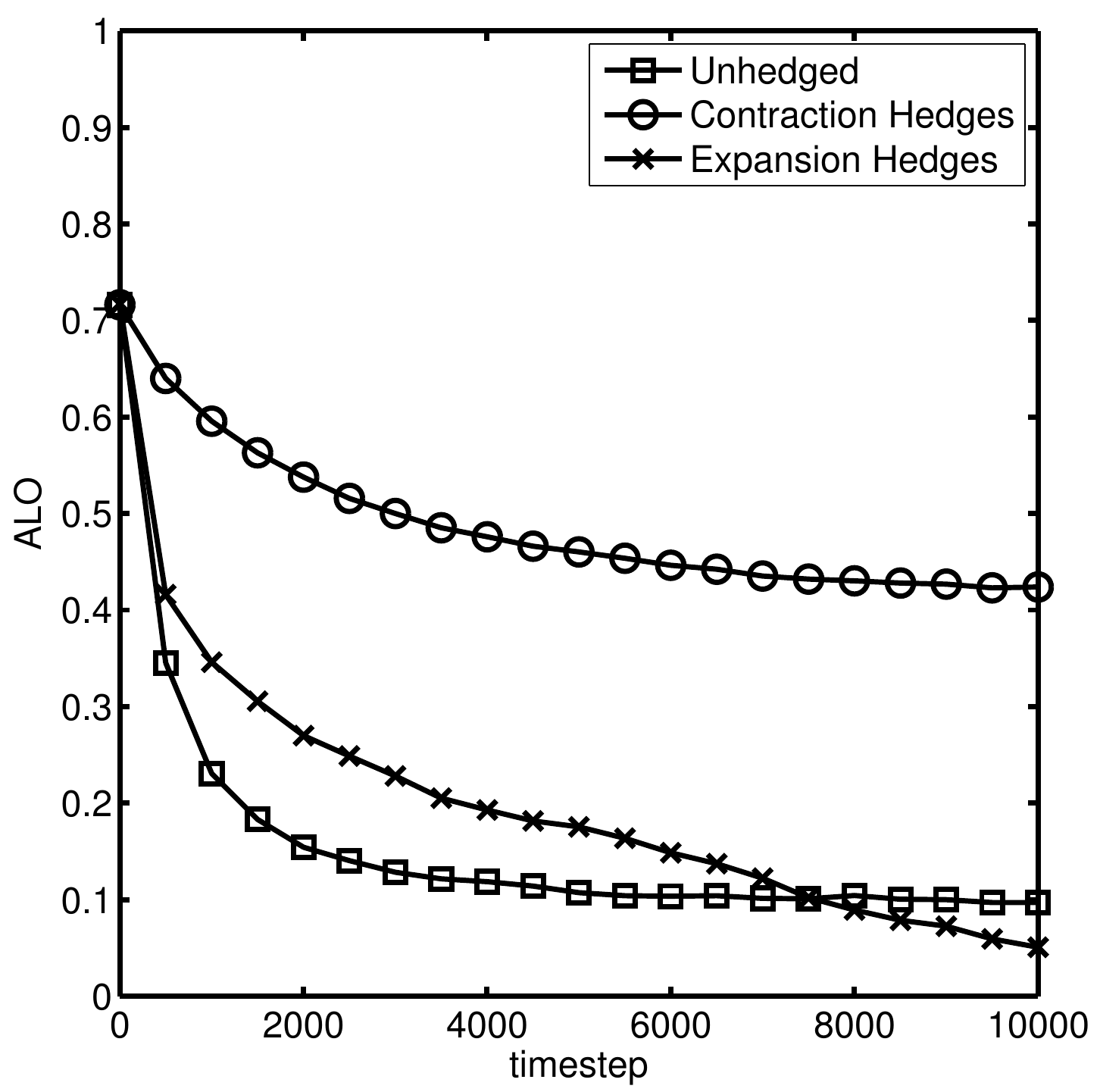}
\caption{ALO vs time for models with no hedged assertions, contracted assertions and expanded assertions, for a prior probability of positive assertions $pp = 0.5$. Although the final value reached is lower when there is a high probability of using expansion hedges, the community of agents takes longer to reach that value, and may even reach a lower value still. }
\label{fig:ALOP05}
\end{figure}

\begin{figure}[h]
\centering
\includegraphics[width = 0.35\textwidth]{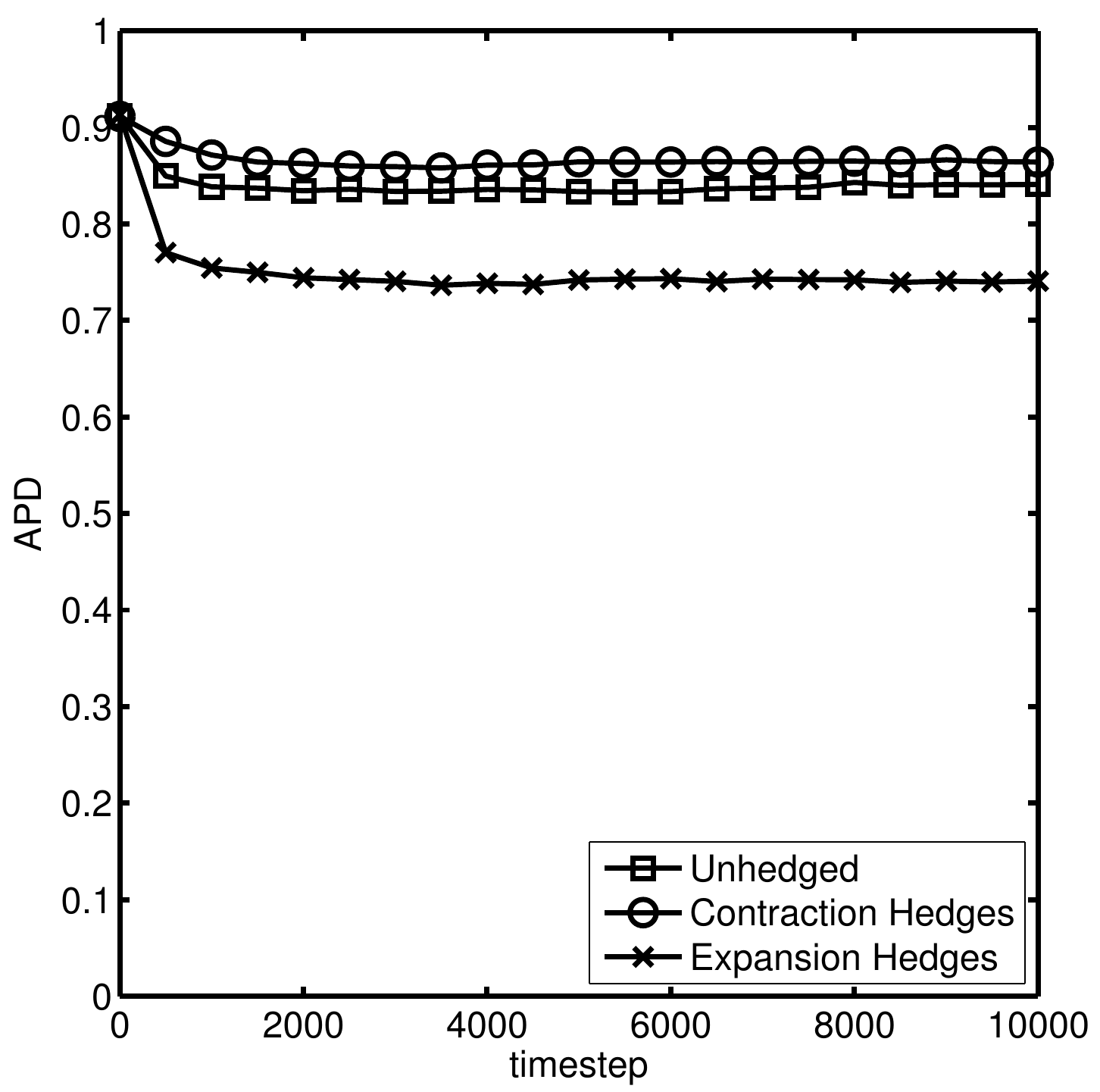}
\caption{APD vs time for models with no hedged assertions, contracted assertions and expanded assertions, for a prior probability of positive assertions $pp = 0.5$. Lower APD is achieved with a high prior probability of asserting expansion hedges, in a similar timescale to the unhedged model}
\label{fig:APDI05}
\end{figure}

We can also examine how fast the community of agents arrives at a steady state. Figure \ref{fig:APDI56} shows that at short timescales ($t < 2000$), better convergence may be achieved allowing only unhedged assertions. In a more extreme case, figure \ref{fig:ALOP05} shows that for $pp = 0.5$, better performance on the ALO metric is only achieved after $7500$ timesteps. Although this improvement takes a longer time to achieve, it goes together with improved performance on APD which is achieved in a similar timescale to the unhedged model \ref{fig:APDI05}.

\section{DISCUSSION}
\label{sec:discussion}
These results show that, in a model of language development across a population, hedged assertions can improve both the level of convergence to shared language as measured by average pairwise difference between label sets (APD) and, to an extent, the discriminatory power of individuals' label sets, as measured by average label overlap (ALO). The two different types of hedges improve performance in distinct ways. If overall convergence is important, a high prior probability of asserting contraction hedges should be used to improve performance on the APD metric. Conversely, if the ability of the agents to discriminate precisely between objects in the environment is more important, then expansion hedges, together with lower probabilities of asserting positive labels, should be used to maintain low levels of ALO whilst still improving performance on APD. 

The improved performance against the two metrics is tempered by the fact that the speed at which the steady state is achieved is somewhat slower than when using simply unhedged assertions. However, the improvement in APD is seen relatively quickly at $pp = 0.56$, soon after the unhedged model has reached its steady state. The improvement in ALO when using expansion hedges, for $pp = 0.5$, does not occur until after $7,500$ timesteps, well after the unhedged model  has reached its steady state. However, the improvement in performance on ALO goes together with improved performance on APD which is attained at the same speed as the in the unhedged model.

If the speed of development of shared categories is not important, the two types of hedges would be useful in different types of situation, depending whether convergence or discriminatory power is more important. This might be dependent on, for example, the structure of the underlying environment. In the current simulation, objects are presented uniformly across the space. If objects were distributed non-uniformly, perhaps clumping in various regions of the space, then perhaps the ability to discriminate precisely between different labels would be less important, since the environment provides that distinction naturally. Convergence to shared labels would then be more important.

If speed is important, using contraction hedges can still improve levels of convergence in a relatively short timeframe.

There are many parameters in the simulation that bear further investigation. The distribution of objects in the environment, as mentioned above, is likely to have an effect on performance again the two metrics. In the current simulations, hedge values of $v = 0.5$ and $h = 2$ are used. Increasing and decreasing these values could have an impact on performance, as would, perhaps, allowing agents to have difference values of $v$ and $h$. The range of $w$ allowed also affects performance. When $w = [0.01, 0.99]$, agents no longer achieve high levels of convergence at $pp > 0.55$ (results not shown). Other weight ranges may positively affect performance, however.

\section{CONCLUSIONS}
\label{sec:conc}
We have investigated the utility of hedged assertions in the development of a shared language, and shown that allowing agents to make hedged assertions improves the ability to develop common categories in two distinct ways. Firstly, using contraction hedges, i.e. words like `very', allows improved levels of convergence to shared categories, whilst slightly improving the extent to which labels overlap. Secondly, using expansion hedges, or words like `quite', enables the development of label sets that are more discriminatory of the environment and also have better levels of convergence. However, both these improvements come with a slower speed of development of shared labels. It may be possible to improve these speeds by tuning other parameters such as the age range of agents or the values of hedges used.

\ack
Martha Lewis gratefully acknowledges support from EPSRC Grant No. EP/E501214/1

\bibliography{../phd}

\end{document}